\pdfoutput=1

\documentclass[11pt]{article}

\usepackage[final]{acl}

\usepackage{times}
\usepackage{latexsym}
\usepackage[T1]{fontenc}

\usepackage[utf8]{inputenc}

\usepackage{microtype}

\usepackage{inconsolata}

\usepackage{graphicx}

\usepackage{amsmath}
\usepackage{algorithm,algorithmic}
\usepackage{amssymb}

\newcommand{\SimLens}{SimLens}
\newcommand{\LinearSimLens}{Linear SimLens}
\newcommand{\SimExit}{SimExit}
\newcommand{\stok}{\ensuremath{\langle s\rangle}}
\newcommand{\atok}{\ensuremath{\langle a\rangle}}

\title{SimLens: Two-Token Continuation for Early Exit in Single-Token Decision Tasks}

\author{
  \textbf{Ming Ma\textsuperscript{1,2,*}},
  \textbf{Bowen Zheng\textsuperscript{3,*}},
  \textbf{Zhongqiao Lin\textsuperscript{1}},
  \textbf{Tianming Yang\textsuperscript{1,\dag}} \\
  \textsuperscript{1}Institute of Neuroscience, \\
  ~~State Key Laboratory of Brain Cognition and Brain-inspired Intelligence Technology, \\
  ~~Center for Excellence in Brain Science and Intelligence Technology, \\
  ~~Chinese Academy of Sciences, Shanghai, 200031, China \\
  \textsuperscript{2}School of Future Technology, University of Chinese Academy of Sciences, \\
  ~~Beijing, 100049, China \\
  \textsuperscript{3}School of Data Science, The Chinese University of Hong Kong, Shenzhen \\
  \texttt{\{mam2022, zqlin, tyang\}@ion.ac.cn} \\
  \texttt{bowenzheng3@link.cuhk.edu.cn} \\
  {\small \textsuperscript{*}These authors contributed equally to this work. \quad \textsuperscript{\dag}Corresponding author}
}

\begin{document}
\maketitle

\begin{abstract}
Intermediate-layer representations in decoder-only large language models (LLMs) contain task-relevant information, yet direct linear readouts often fail to recover the model's eventual choice. This gap limits both interpretability and early-exit methods for candidate-scoring workloads. We study \emph{single-token decision} settings in which a small candidate set is supplied by the task format, such as multiple-choice and boolean QA. We propose \SimLens{}, a training-free readout that, at a chosen intermediate layer, keeps only the start token and the candidate answer token and forwards this reduced two-token state through the model's remaining upper layers. To identify the layer at which the full sequence can be safely truncated, we distill \LinearSimLens{} as a lightweight entropy monitor and combine the two in \SimExit{}; unlike \SimLens{}, this monitor is trained. Across five decoder-only backbones from 3B to 8B parameters, \SimLens{} produces earlier and more faithful candidate readout than direct linear lenses, while \SimExit{} delivers a modest but stable speed--accuracy trade-off on the two models used end-to-end under fair-compute accounting. Ablations show that the two retained tokens play complementary roles as global condition and semantic anchor.
\end{abstract}

\section{Introduction}
\begin{figure}[!t]
    \centering
    \includegraphics[width=0.96\linewidth]{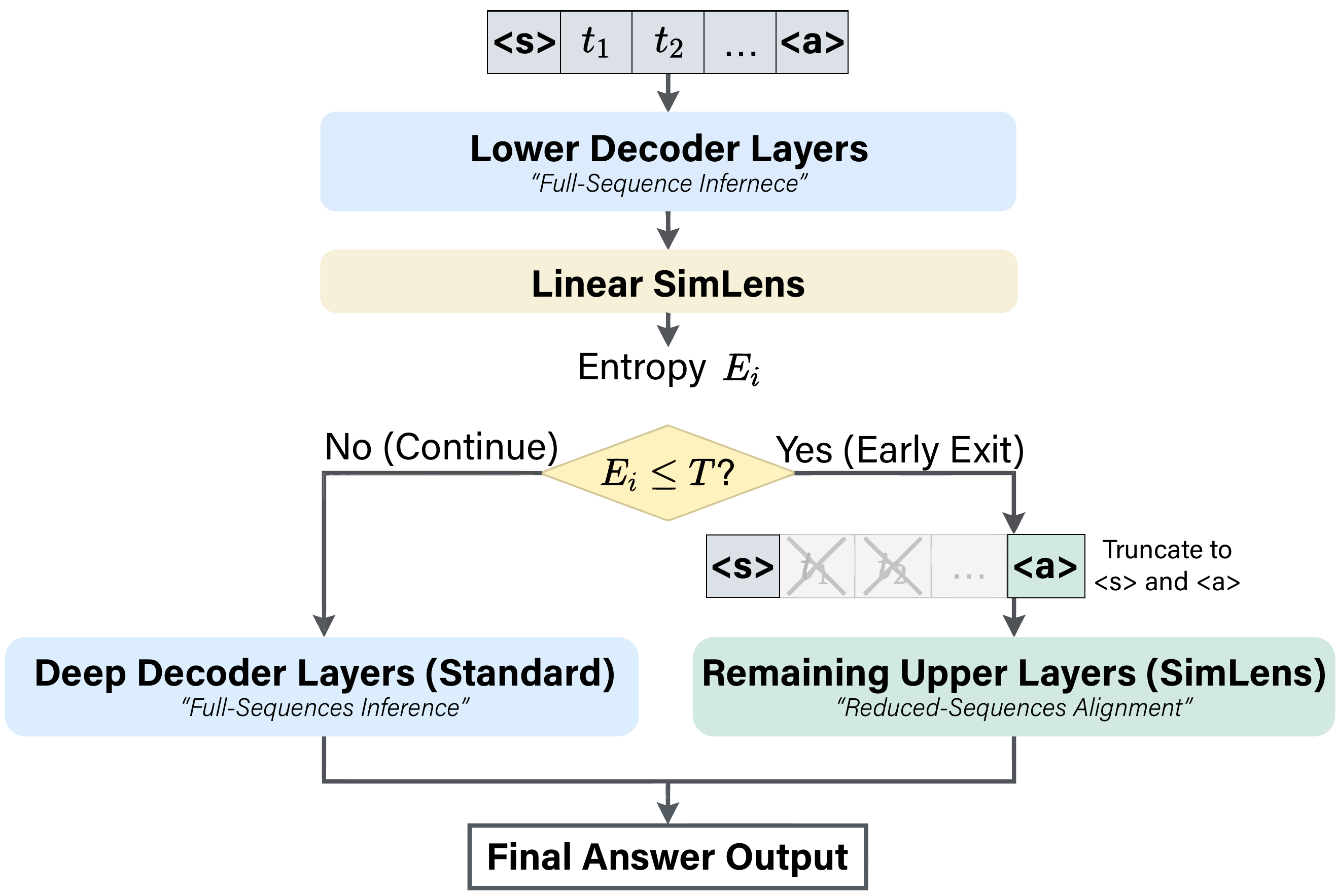}
    \caption{Overview of \SimExit{}. At each monitored layer, \LinearSimLens{} provides a low-cost entropy confidence score for exit decisions. Once the confidence threshold is met, full-sequence forwarding is truncated and \SimLens{} continues with only two tokens---the start token and the candidate answer token---to decode the answer.}
    \label{fig:method}
\end{figure}

Large language models (LLMs) form their predictions gradually across depth. If we could accurately read out a model's latent prediction from an intermediate layer, this would help us understand how semantics emerge inside the network and could also reduce inference cost. Motivated by this, prior work studies hidden-state decoding with lens-style methods (\citealp{interpreting}; \citealp{dar2023analyzing}; \citealp{geva2023dissecting}; \citealp{hendel2023incontext}; \citealp{belrose2023eliciting}; \citealp{ghandeharioun2024patchscopes}; \citealp{skean2024does}; \citealp{skean2025layer}), and many early-exit methods attempt to stop computation at intermediate layers to reduce cost (\citealp{schuster2022confident}; \citealp{pal2023future}; \citealp{din2024jump}; \citealp{valade2024accelerating}; \citealp{elhoushi2024layerskip}; \citealp{fan2024not}; \citealp{zhang2024draft}; \citealp{kavehzadeh2024sorted}; \citealp{jazbec2024fast}; \citealp{miao2024efficient}; \citealp{mofakhami2024performance}; \citealp{fan2025position}; \citealp{zarch2025context}).

Existing approaches such as Logit Lens, Tuned Lens, and Future Lens read intermediate states with shared or learned linear projections. These methods are useful analysis tools, but decoding from earlier layers remains unreliable: even when answer-relevant information is already present, the decoded output can still deviate sharply from the model's final prediction. On ARC \citep{clark2018thinksolvedquestionanswering}, for example, a linear probe recovers answer information much earlier than Logit Lens \citep{interpreting}, suggesting that the bottleneck is not the absence of latent information, but the difficulty of reading it out faithfully. We provide the corresponding motivation visualizations in Appendix~\ref{app:motivation}.

We focus on a specific regime: \emph{candidate-scoring} settings where the candidate label or answer token is provided by the task format. This includes multiple-choice QA, boolean verification, reranking, routing, moderation, and preference/reward scoring, and differs from unrestricted long-form generation. \SimLens{} reads out intermediate candidate scores in this scoped setting; \SimExit{} uses this readout for early exit. We do not claim \SimLens{} as a general-purpose accelerator for long-form generation. For example, a RAG verifier may repeatedly decide whether a candidate answer is supported by retrieved evidence using a fixed \texttt{Yes}/\texttt{No} output; \SimExit{} can stop full-sequence processing once that decision is confident and continue only the two retained tokens.

Within this scope we take a different route from stronger linear lenses. We ask whether a \emph{minimal two-token continuation} can recover a much more faithful intermediate-layer prediction. We propose \textbf{\SimLens{}}, a training-free readout that, at an intermediate layer $l$, retains only the start token and the candidate answer token---written $\langle s \rangle$ and $\langle a \rangle$ in equations---and forwards this two-token state through the remaining upper layers $T_{l+1},\dots,T_L$. The two tokens play complementary roles: the start token preserves global conditioning, while the candidate answer token serves as a semantic anchor for the candidate being scored.

This minimal continuation is already enough to outperform linear baselines at early layers and often moves much closer to the final model prediction. We evaluate five decoder-only 3--8B backbones---LLaMA-7B, Vicuna-7B, Llama-3.1-8B, Qwen2.5-7B-Instruct, and Qwen2.5-3B-Instruct---on ARC \citep{clark2018thinksolvedquestionanswering}, BoolQ \citep{clark2019boolqexploringsurprisingdifficulty}, and HeadQA \citep{vilares2019headqahealthcaredatasetcomplex}. In this setting the candidate answer token is given by the task format itself, so \SimLens{} scores each candidate directly without needing a separate full-model decode to first discover the answer token.

More faithful intermediate readout naturally enables early exit. We therefore introduce \textbf{\LinearSimLens{}}, a lightweight linear monitor distilled from \SimLens{} for low-cost entropy estimation, and combine it with \SimLens{} in \textbf{\SimExit{}}, a hybrid early-exit framework (Figure~\ref{fig:method}). \SimLens{} itself is training-free; \LinearSimLens{} and \SimExit{} use a small trained monitor. Across LLaMA-7B \citep{touvron2023llama} and Vicuna-7B \citep{vicuna2023}, \SimLens{} improves iso-compute accuracy over matched-cost naive intermediate decoding in all six model-dataset settings, with an average gain of $+43.2$\,pp after charging the extra two-token post-forward cost. \SimExit{} yields an average $1.15\times$ speedup at the best-accuracy operating points and $1.40\times$ when allowing up to a 1\,pp accuracy drop.

\paragraph{Contributions.} We make three contributions: (1) we introduce \SimLens{}, a training-free readout that uses a two-token continuation to elicit accurate candidate predictions from intermediate layers of LLMs; (2) we show that retaining the start token and the candidate answer token is sufficient to markedly improve early-layer readout, and we clarify their distinct roles as global condition and semantic anchor; and (3) we introduce \LinearSimLens{}, a lightweight distilled monitor, and \SimExit{}, a hybrid early-exit framework that combines low-cost entropy monitoring with the two-token \SimLens{} readout.

\begin{figure*}[!t]
    \centering
    \includegraphics[width=\linewidth]{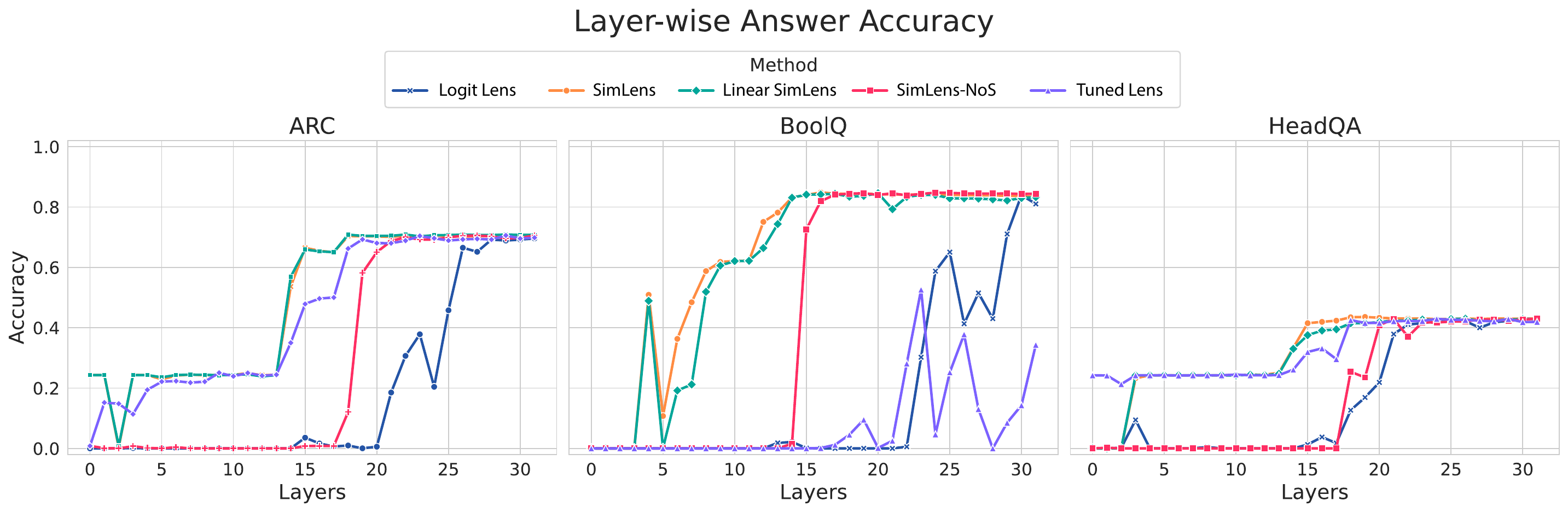}
    \caption{Layer-wise answer accuracy on ARC (left), BoolQ (middle), and HeadQA (right). We compare Logit Lens, Tuned Lens, \SimLens{}, \LinearSimLens{}, and \SimLens{}-NoS (\SimLens{} without the start token). Dashed lines show naive full-model accuracy. \SimLens{} variants decode useful answers earlier than linear baselines, while removing the start token consistently reduces performance.}
    \label{fig:performance_improvement}
\end{figure*}

\section{Related Work}

\paragraph{Hidden representation decoding.}
Intermediate layers in LLMs encode substantial task-relevant semantics (\citealp{hendel2023incontext,wang2023label,halawi2024overthinking,din2024jump,skean2024does,skean2025layer}). Logit Lens reuses the output embedding to map each hidden state to logits \citep{interpreting}; Tuned Lens \citep{belrose2023eliciting} and Future Lens \citep{pal2023future} add learned linear projections; PatchScope is a training-free variant \citep{ghandeharioun2024patchscopes}. Activation patching shows that many tasks survive with most sequential context removed when intervention is performed in middle layers (\citealp{hendel2023incontext,wang2023label,pal2023future,zheng2024distributed}). Our work is closest to learned-lens methods but asks a different question: in single-token decision settings, can we replace a stronger linear projector with the model's own upper layers applied to a reduced two-token sequence? \SimLens{} is therefore a sequence-axis alternative to direct linear readout rather than another lens variant.

\paragraph{Early exit.}
Early exit is one of several directions for reducing LLM inference cost, alongside quantization (\citealp{zhang2024flattenquant}; \citealp{hu2024llm}; \citealp{hasan2024optimizing}), pruning (\citealp{li2024greedy}; \citealp{fang2024maskllm}), and distillation (\citealp{wang2024feature}; \citealp{ko2024distillm}; \citealp{di2024performance}). Early-exit, layer-skip, and self-speculative methods---threshold-based exits (\citealp{schuster2022confident}; \citealp{zhang2024draft}; \citealp{fan2024not}; \citealp{jazbec2024fast}; \citealp{mofakhami2024performance}), architecture-based exits (\citealp{pal2023future}; \citealp{din2024jump}; \citealp{valade2024accelerating}; \citealp{kavehzadeh2024sorted}), and LayerSkip~\citep{elhoushi2024layerskip}, D3~\citep{fan2025position}, and DEL~\citep{zarch2025context}---reduce computation along the \emph{depth axis}: they cut the layers run per token to accelerate long-form generation while preserving full sequence length at every layer, with system support for the resulting KV-cache management \citep{miao2024efficient}. \SimExit{} instead reduces along the \emph{sequence axis}: after exiting, it forwards only the two-token reduced state, and it targets single-token decisions rather than long-form generation. The two families therefore optimize different axes for different workloads---depth-wise acceleration for generation, \SimExit{} for candidate scoring.

\section{Preliminaries}
\label{sec:preliminaries}
We consider a language model $\mathcal{M}$ with $L$ layers, indexed by $l \in \{1,\dots,L\}$. For an input sequence $S=\{x_1,\dots,x_n\}$ the input hidden states are $H^0 \in \mathbb{R}^{n \times d}$, with rows $h_i^0 = E(x_i)$ given by the token embedding $E(\cdot)$. We write the $l$-th transformer block as $T_l$, a sequence-level operator combining causal self-attention and a feed-forward sublayer:
\begin{equation}
\label{eq:transformer_layer}
    H^l = T_l(H^{l-1}), \quad l=1,\dots,L.
\end{equation}
Let $h_i^l$ denote the $i$-th row of $H^l$.

At the final layer $L$, a linear output projection maps each hidden state to vocabulary logits:
\begin{equation}
\label{logits equation}
    z_i^L = W h_i^L,
\end{equation}
where $W \in \mathbb{R}^{V \times d}$ is the output projection matrix. The token distribution is then
\begin{equation}
\label{softmax equation}
    p_i = \mathrm{softmax}(z_i^L)
\end{equation}
with $p_i \in \mathbb{R}^{V}$.

\begin{table*}[t]
\centering
\resizebox{\textwidth}{!}{
\begin{tabular}{l l c c c c}
\hline
Model & Dataset & Best SimLens Layer & SimLens Acc.\ (\%) & Matched-Layer Naive Acc.\ (\%) & $\Delta$ vs.\ Naive (pp) \\
\hline
LLaMA-7B & ARC & L24 & 70.6 & 20.4 & +50.2 \\
LLaMA-7B & BoolQ & L16 & 84.9 & 0.0 & +84.9 \\
LLaMA-7B & HeadQA & L19 & 43.6 & 16.9 & +26.7 \\
Vicuna-7B & ARC & L26 & 75.4 & 72.9 & +2.5 \\
Vicuna-7B & BoolQ & L21 & 84.6 & 10.1 & +74.5 \\
Vicuna-7B & HeadQA & L20 & 51.2 & 30.6 & +20.6 \\
\hline
Mean & -- & L21 & 68.4 & 25.2 & +43.2 \\
\hline
\end{tabular}
}
\caption{Layer-wise readout summary at the \SimLens{} best layer. For each model-dataset pair, we report the best \SimLens{} layer, \SimLens{} accuracy at that layer, the matched-layer naive intermediate decoding accuracy, and their difference in percentage points (pp).}
\label{tab:key_layer_acc}
\end{table*}

\begin{table*}[t]
\centering
\resizebox{\textwidth}{!}{
\begin{tabular}{l l c c c c}
\hline
Model & Dataset & Full-Depth Acc.\ (\%) & Best Logit Lens & Best SimLens & $\Delta$ (pp) \\
\hline
Llama-3.1-8B & ARC & 83.5 & L31 / 83.9 & L26 / 82.8 & $-1.1$ \\
Llama-3.1-8B & BoolQ & 63.0 & L29 / 71.9 & L18 / 72.1 & $+0.2$ \\
Qwen2.5-7B-Instruct & ARC & 92.8 & L28 / 92.8 & L27 / 92.9 & $+0.1$ \\
Qwen2.5-7B-Instruct & BoolQ & 85.9 & L27 / 86.3 & L28 / 86.9 & $+0.6$ \\
Qwen2.5-3B-Instruct & ARC & 90.8 & L35 / 90.3 & L31 / 90.9 & $+0.6$ \\
Qwen2.5-3B-Instruct & BoolQ & 83.7 & L34 / 83.5 & L28 / 84.2 & $+0.7$ \\
\hline
\end{tabular}
}
\caption{Best-layer readout on additional 3B--8B backbones under the candidate-scoring protocol. Best-layer differences against Logit Lens are small (between $-1.1$ and $+0.7$\,pp); Qwen2.5-3B-Instruct was added during the rebuttal period.}
\label{tab:additional_backbones}
\end{table*}

\section{SimLens}
\label{sec:simlens}

\SimLens{} is a training-free readout that reuses the model's native forward dynamics to elicit a prediction from an intermediate layer, without any auxiliary training. In this paper we instantiate it on \emph{single-token decision} tasks, in which the answer is a single token at a position fixed by the task format---an answer-choice letter in multiple-choice QA (e.g., ARC and HeadQA) or a yes/no token in boolean QA (e.g., BoolQ). Because this answer position is known from the prompt, \SimLens{} reads out the model's next-token prediction there directly at an intermediate layer---its highest-probability token over the full vocabulary---without first running a full forward pass to obtain the answer.

For a chosen intermediate layer $l$, \SimLens{} keeps only two tokens: the start token \stok{} and the candidate answer token \atok{} (the final prompt token, whose next-token prediction gives the answer). Let $h^l_{\stok{}}$ and $h^l_{\atok{}}$ be their layer-$l$ hidden states, and let $T_{l+1:L} = T_L \circ T_{L-1} \circ \cdots \circ T_{l+1}$ denote the composition of the remaining upper blocks. \SimLens{} forwards only these two hidden states through the upper blocks:
\begin{equation}
\label{eq:simlens}
[\, h^L_{\stok{}},\ h^L_{\atok{}} \,] = T_{l+1:L}\big([\, h^l_{\stok{}},\ h^l_{\atok{}} \,]\big),
\end{equation}
where $[\cdot,\cdot]$ denotes the ordered two-token state. We then convert $h^L_{\atok{}}$, the upper-layer representation of the candidate answer token, to a distribution over the full vocabulary via Eq.~(\ref{logits equation}) and Eq.~(\ref{softmax equation}), and take its highest-probability token as the model's prediction.

The two retained tokens play complementary roles. The start token provides global conditioning and a stable context anchor, consistent with recent analyses of first-token attention and attention-sink behavior in LLMs \citep{gu2024attention,barbero2025llms}; the candidate answer token provides a semantic anchor for the answer prediction at that position. This two-token continuation is the core design of \SimLens{}: it adds very little compute, yet preserves enough upper-layer non-linear processing to recover substantially more faithful latent predictions than a direct linear readout.

\subsection{Linear SimLens: Approximation With Linear Projection}
\label{sec:linear_simlens}
\SimLens{} is accurate but still requires a two-token forward pass from layer $l$ to layer $L$. For frequent confidence checks inside an early-exit loop we approximate \SimLens{} with a lightweight linear mapping, termed \emph{\LinearSimLens{}}, which is trained.

For a monitored layer $l$ with $0 < l \le L$, we learn a linear map $\mathcal{F}_l$ that predicts the final-layer representation from the layer-$l$ hidden state:
\begin{equation}
    \hat{h}^L_i = \mathcal{F}_l(h^l_i),
\end{equation}
yielding predicted logits $\hat{z}^L_i = W \hat{h}^L_i$. Unlike Tuned Lens \citep{belrose2023eliciting}, the distillation targets are not produced by the original model: the target logits $z^L_i$ are the \SimLens{} logits from Eq.~(\ref{eq:simlens}) at layer $l$. With temperature-scaled softmax distributions $p^{SL}_l = \mathrm{softmax}(z^L_i / \tau)$ (the \SimLens{} target) and $\hat{p}_l = \mathrm{softmax}(\hat{z}^L_i / \tau)$ (the \LinearSimLens{} prediction), where $\tau$ is a distillation temperature, we train $\mathcal{F}_l$ by minimising the cross-entropy between them
\begin{equation}
\label{eq:linear_simlens_loss}
\mathcal{L}_l
= - \!\! \sum_{v \in V} p^{SL}_{l,v} \log \hat{p}_{l,v}.
\end{equation}
By distilling from \SimLens{} rather than from the original model, \LinearSimLens{} focuses on the representational shift introduced by the two-token continuation while avoiding any full-sequence propagation during training and inference-time monitoring. Full hyperparameters are given in Appendix~\ref{app:linear_simlens_repro}.

\subsection{Setup}
\label{sec:setup}
We test LLaMA-7B \citep{touvron2023llama}, Vicuna-7B \citep{vicuna2023}, Llama-3.1-8B \citep{grattafiori2024llama3}, Qwen2.5-7B-Instruct, and Qwen2.5-3B-Instruct \citep{yang2024qwen25}. Single-token decision benchmarks are ARC \citep{clark2018thinksolvedquestionanswering}, BoolQ \citep{clark2019boolqexploringsurprisingdifficulty}, and HeadQA \citep{vilares2019headqahealthcaredatasetcomplex}. All five backbones use the same downstream data, code, and LoRA adaptation workflow \citep{hu2022lora}; hence the cross-generation comparison is not confounded by fine-tuning status. We report layer-wise readout accuracy for all five models, while LLaMA-7B and Vicuna-7B drive the full fair-compute and \SimExit{} evaluation.

\subsection{Layer-wise Readout Accuracy}
\label{sec:exp_readout}
Figure~\ref{fig:performance_improvement} compares layer-wise accuracy on LLaMA-7B and Vicuna-7B across Logit Lens, Tuned Lens, \SimLens{}, \LinearSimLens{}, and \SimLens{}-NoS on ARC, BoolQ, and HeadQA. Both \SimLens{} and \LinearSimLens{} decode useful answers at earlier layers than the linear baselines and reach the naive model's accuracy at shallower depths. On ARC, \SimLens{} rises above the random baseline at layer 14 and reaches the naive LLM's accuracy by layer 18, while Logit Lens only starts to rise at layer 21 and remains below \SimLens{} until layer 28; the same earlier-readout trend appears on BoolQ and HeadQA. Removing the start token (\SimLens{}-NoS) substantially lowers ARC readout, which only reaches naive LLM accuracy after layer 20. Table~\ref{tab:key_layer_acc} quantifies the gap at the \SimLens{} best layer: across the six LLaMA-7B and Vicuna-7B settings, \SimLens{} reaches mean accuracy $68.4\%$ compared to $25.2\%$ for naive intermediate decoding at the same layer---an average gain of $+43.2$\,pp.

Tuned Lens, trained on ARC, is competitive in-domain (70.60\% at layer 29) but remains slightly below \LinearSimLens{} (70.91\%); its cross-dataset transfer is less stable, with best accuracies of 52.63\% on BoolQ and 42.92\% on HeadQA, both well below \SimLens{} (84.86\% and 43.58\%) and \LinearSimLens{} (84.59\%). To test cross-task transfer of \LinearSimLens{}, we train it once on ARC and apply it directly to BoolQ, HeadQA, and WikiText-103 \citep{merity2016pointer}; the layer-wise cross-entropy curves (Figure~\ref{fig:simlens_compare} in Appendix~\ref{app:ce_curves}) show training-free \SimLens{} attains the lowest CE across layers, while \LinearSimLens{} remains competitive at much lower monitoring overhead. This supports the motivation that improving the linear readout alone is not enough to replace upper-layer non-linearity, especially under cross-dataset transfer.

The same earlier-readout pattern is reproduced on Llama-3.1-8B, Qwen2.5-7B-Instruct, and Qwen2.5-3B-Instruct under the candidate-scoring protocol (Table~\ref{tab:additional_backbones}). On ARC, \SimLens{} improves mid-depth readout by $+14.8$\,pp at L16 on Llama-3.1-8B and by $+39.6$\,pp at L21 on Qwen2.5-7B-Instruct; on BoolQ, it reaches full-depth candidate accuracy three to four layers earlier than Logit Lens. On Qwen2.5-3B-Instruct, the best \SimLens{} readout appears four layers earlier on ARC and six layers earlier on BoolQ, with best-layer gains of $+0.6$ and $+0.7$\,pp. Thus, across the evaluated 3B--8B models, the supported claim is earlier faithful readout rather than a universal scaling law or uniformly higher best-layer accuracy. Full 7B/8B layer-wise curves are in Appendix~\ref{app:additional_backbones}.

\subsection{Fair-Compute Analysis}
\label{sec:exp_fair}
\begin{figure*}[t]
    \centering
    \includegraphics[width=0.95\linewidth]{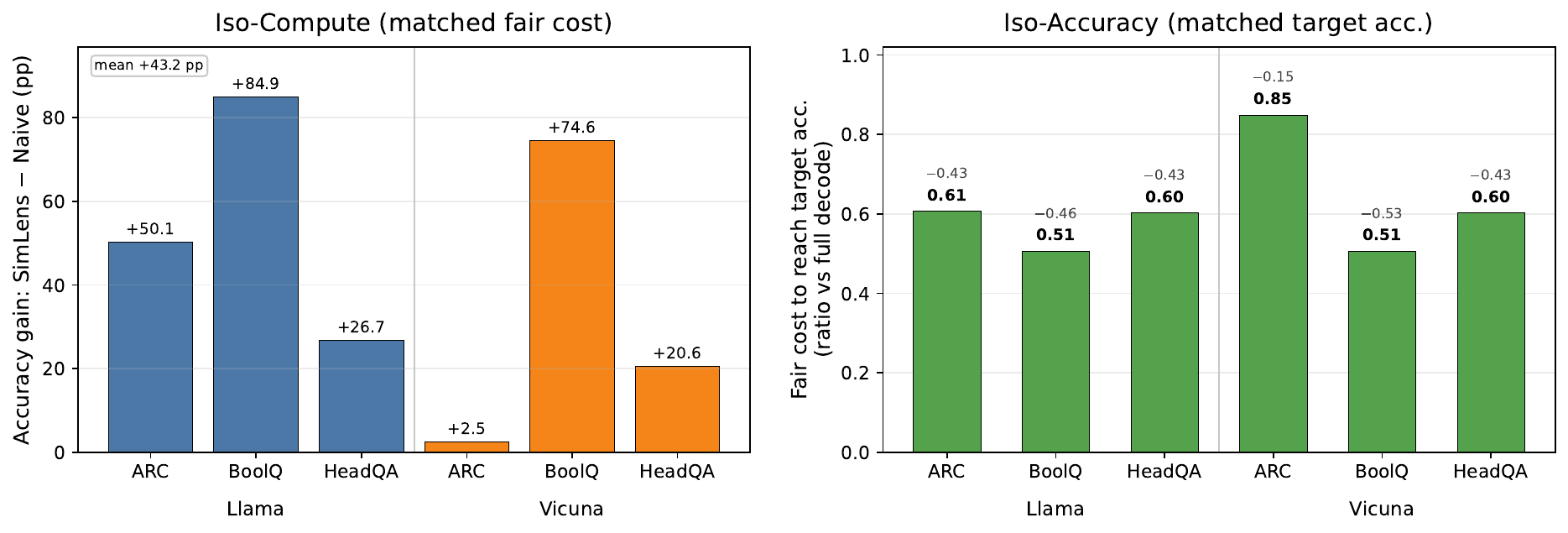}
    \caption{Iso-compute and iso-accuracy alignment under fair-cost accounting. \textbf{Left:} accuracy gain of \SimLens{} over matched-cost naive decoding at iso-compute (pp). \textbf{Right:} \SimLens{}'s fair cost to reach the matched-layer target accuracy, as a ratio to full-depth decoding; the gray number above each bar is the fair-cost difference vs.\ naive decoding.}
    \label{fig:iso_compute_alignment}
\end{figure*}

The two-token continuation adds a small post-forward cost beyond the matched-layer naive decoding. To make the cost comparison explicit, we use two complementary criteria. \emph{Iso-compute} compares methods at matched fair cost, while \emph{iso-accuracy} reports the fair cost required by \SimLens{} to reach the matched-layer target accuracy. In both cases, fair cost includes the two-token post-forward overhead. Table~\ref{tab:iso_compute} and Figure~\ref{fig:iso_compute_alignment} re-express this gain under fair-cost accounting. At matched compute, \SimLens{} stays more accurate than naive decoding in all six LLaMA-7B/Vicuna-7B settings; this iso-compute advantage is the matched-layer gap already reported in Table~\ref{tab:key_layer_acc} (mean $+43.2$\,pp), now charging \SimLens{} its two-token overhead (mean fair cost $0.693$ vs.\ $0.688$ for naive at matched compute). Under iso-accuracy, \SimLens{} reaches the matched-layer target accuracy with mean fair cost $0.612$, well below its best-layer cost.

\begin{table*}[!t]
\centering
\resizebox{\textwidth}{!}{
\begin{tabular}{l l c c c c}
\hline
Model & Dataset & SimLens Fair Cost & Naive Cost at Iso-Compute & $\Delta$ Acc.\ at Iso-Compute (pp) & SimLens Cost at Iso-Accuracy \\
\hline
LLaMA-7B & ARC & 0.788 & 0.781 & $+50.2$ & 0.606 \\
LLaMA-7B & BoolQ & 0.537 & 0.531 & $+84.9$ & 0.506 \\
LLaMA-7B & HeadQA & 0.633 & 0.625 & $+26.7$ & 0.602 \\
Vicuna-7B & ARC & 0.849 & 0.844 & $+2.5$ & 0.849 \\
Vicuna-7B & BoolQ & 0.691 & 0.688 & $+74.5$ & 0.506 \\
Vicuna-7B & HeadQA & 0.663 & 0.656 & $+20.6$ & 0.602 \\
\hline
Mean & -- & 0.693 & 0.688 & $+43.2$ & 0.612 \\
\hline
\end{tabular}
}
\caption{Iso-compute and iso-accuracy alignment under fair-cost accounting. Fair cost includes the extra two-token post-forward overhead of \SimLens{}. \SimLens{} improves iso-compute accuracy in all six settings and reaches the matched-layer target accuracy with substantially lower fair cost.}
\label{tab:iso_compute}
\end{table*}

\section{SimExit: Accelerating LLMs with SimLens}
\label{sec:simexit}

\LinearSimLens{} offers cheap decoding with a single linear map; \SimLens{} remains more accurate, especially under cross-task transfer. We combine the two in \SimExit{}, a hybrid early-exit framework that uses \LinearSimLens{} as a fast confidence monitor and \SimLens{} as the final readout.

\subsection{Entropy-Based Confidence with Linear SimLens}
Because confidence must be re-evaluated repeatedly during the forward pass, we use \LinearSimLens{} for monitoring. We use the entropy of the \LinearSimLens{} distribution as the exit signal:
\begin{equation}
  E^{(l)} \;=\; -\sum_{v=1}^{V} p^{(l)}_{v}\,\log p^{(l)}_{v},
\end{equation}
where $p^{(l)}\!\in\!\mathbb{R}^{V}$ is the \LinearSimLens{} softmax distribution at layer $l$. Lower $E^{(l)}$ indicates more concentrated probability mass and higher confidence. Entropy is less sensitive than the top-2 gap (\citealp{fan2024not,din2024jump,jazbec2024fast,mofakhami2024performance}) to ties among several plausible tokens, especially in settings where multiple candidates remain competitive.

\subsection{Adaptive Early-Exit Mechanism}
When entropy falls below a fixed threshold $\tau_{\mathrm{exit}}$, the full-sequence forward pass is truncated and \SimLens{} completes decoding with only the start and candidate answer tokens, using the upper transformer blocks $T_{l+1:L}$ in Eq.~(\ref{eq:simlens}). Similar thresholding mechanisms appear in biological decision-making (\citealp{gold2002banburismus,gold2007neural,kira2015neural}); tuning $\tau_{\mathrm{exit}}$ trades accuracy for speed.

To control monitoring overhead, entropy is checked every $N$ layers ($l \bmod N = 0$). A smaller $N$ checks confidence more frequently and can trigger earlier exits, but adds more \LinearSimLens{} evaluations; a larger $N$ reduces monitoring cost but may delay truncation. We select $N$ on a validation set by maximizing speedup under an accuracy-drop constraint $\delta$:
\begin{equation}
\label{eq:n_selection}
\begin{aligned}
N^{\star}=\arg\max_{N}\;&\text{Speedup}(N, \tau_{\mathrm{exit}})\\
\text{s.t.}\;&\Delta\text{Acc}(N, \tau_{\mathrm{exit}})\le\delta.
\end{aligned}
\end{equation}

With the original LLM, a full forward pass over sequence length $n$ has $\mathcal{O}(n^2)$ attention cost. After exit, \SimExit{} forwards only two tokens, so the post-exit cost is effectively independent of $n$. The complete procedure is given in Algorithm~\ref{alg:adaptive_inference}.

\begin{figure*}[htb]
    \centering
    \includegraphics[width=0.95\linewidth]{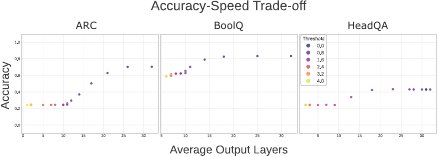}
    \caption{\SimExit{} performance on ARC (left), BoolQ (middle), and HeadQA (right). Colors indicate different entropy thresholds $\tau_{\mathrm{exit}}$. Lower thresholds delay exit and preserve accuracy; higher thresholds exit earlier and trade accuracy for speed.}
    \label{fig:accelerate_llama}
\end{figure*}

\begin{figure*}[htb]
    \centering
    \includegraphics[width=0.95\linewidth]{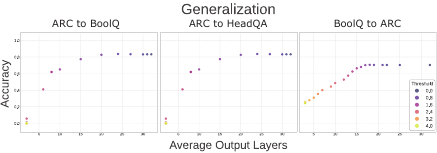}
    \caption{Cross-task transfer for \SimExit{}. Left/middle: evaluate on BoolQ and HeadQA using \LinearSimLens{} trained on ARC. Right: evaluate on ARC using \LinearSimLens{} trained on BoolQ. The method preserves a favorable speed--accuracy trade-off without per-task retraining.}
    \label{fig:generalization}
\end{figure*}

\subsection{Performance}
\label{sec:exp_simexit}
We evaluate \SimExit{} on ARC, BoolQ, and HeadQA with LLaMA-7B (Figure~\ref{fig:accelerate_llama}); Vicuna-7B curves are in Appendix~\ref{app:exit_perf}. Sweeping the entropy threshold $\tau_{\mathrm{exit}}$ traces a speed--accuracy frontier within the single-token decision setting. Because \LinearSimLens{} transfers across tasks, an \SimExit{} monitor trained on one task remains competitive on other tasks (Figure~\ref{fig:generalization}).

\subsection{Threshold Sensitivity}
\begin{figure*}[t]
  \centering
  \includegraphics[width=\linewidth]{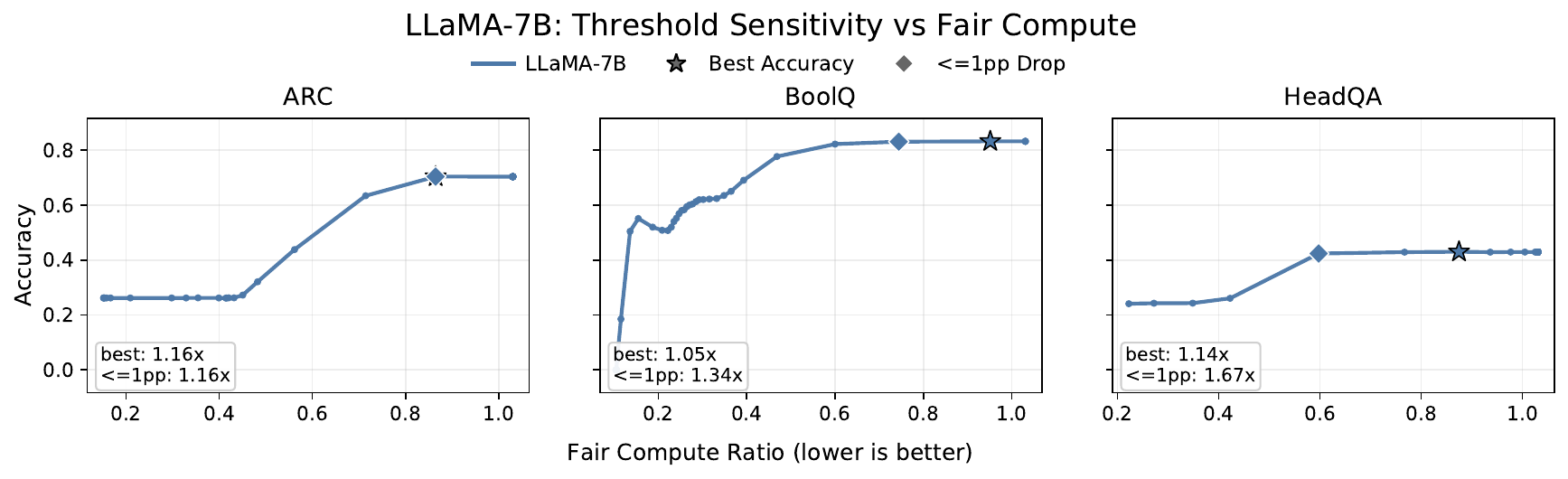}
  \caption{Accuracy vs.\ fair compute under threshold sweeping for LLaMA-7B. Fair compute includes the $+2$-token post-forward overhead after truncation. Star markers indicate best-accuracy operating points; diamond markers indicate operating points within $1$\,pp accuracy drop.}
  \label{fig:threshold_fair_compute_llama}
\end{figure*}

We quantify the threshold trade-off using fair compute, which includes the extra two-token post-forward cost. Averaged over the six settings, \SimExit{} obtains $1.15\times$ speedup at the best-accuracy operating points and $1.40\times$ when allowing up to a $1$\,pp accuracy drop. These gains are modest but are reported under fair accounting and without modifying the frozen LLM. Figure~\ref{fig:threshold_fair_compute_llama} shows the LLaMA-7B fair-compute frontier across the three tasks; the per-setting operating points underlying these speedups are listed in Table~\ref{tab:threshold_sensitivity_extended}, and the Vicuna-7B frontier in Appendix~\ref{app:threshold_tables}. For deployment, a simple policy is to first choose an allowable accuracy drop on a validation set (e.g., at most $1$\,pp), and then select the threshold with the lowest fair cost under that constraint.

\begin{algorithm}[t]
\caption{\SimExit{} (single-token decision)}
\label{alg:adaptive_inference}
\begin{algorithmic}[1]
\REQUIRE Frozen LLM with layers $T_1,\dots,T_L$; pre-trained \LinearSimLens{} maps $\{\mathcal{F}_l\}$; entropy threshold $\tau_{\mathrm{exit}}$; monitoring interval $N$
\REQUIRE Input embeddings $H^0$ containing the start token \stok{} and the candidate answer token \atok{}
\FOR{$l = 1, \dots, L$}
    \STATE $H^l \gets T_l(H^{l-1})$ \hfill\COMMENT{full-sequence forward}
    \IF{$l \bmod N = 0$}
        \STATE $\hat{z}^L \gets W \mathcal{F}_l(h^l_{\atok{}})$ \hfill\COMMENT{\LinearSimLens{} at \atok{}}
        \STATE $E^{(l)} \gets \mathrm{Entropy}(\mathrm{softmax}(\hat{z}^L))$
        \IF{$E^{(l)} \le \tau_{\mathrm{exit}}$}
            \STATE $[\, h^L_{\stok{}},\ h^L_{\atok{}} \,] \gets T_{l+1:L}\big([\, h^l_{\stok{}},\ h^l_{\atok{}} \,]\big)$ \hfill\COMMENT{two-token continuation}
            \STATE \textbf{return} prediction from $h^L_{\atok{}}$ \hfill\COMMENT{\SimLens{} readout}
        \ENDIF
    \ENDIF
\ENDFOR
\STATE \textbf{return} prediction from $h^L_{\atok{}}$ \hfill\COMMENT{no early exit: full-depth readout}
\end{algorithmic}
\end{algorithm}

\section{Ablation Studies}
\label{sec:ablation}

\subsection{Why the Candidate Answer Token Matters: Semantic Anchor}
\label{sec:ablation_a}
We run a wrong-anchor ablation on ARC: keeping the start token fixed, we replace only the candidate answer token \atok{} in the reduced two-token state (Table~\ref{tab:wrong_a}).

\begin{table}[t]
\centering
\small
\begin{tabular}{l r r}
\hline
Variant & Acc.\ (\%) & $\Delta$ vs.\ Correct-A (pp) \\
\hline
Correct-A & 61.4 & 0.0 \\
Wrong-A & 48.2 & $-13.2$ \\
Random-A & 24.9 & $-36.5$ \\
Zero-A & 25.2 & $-36.2$ \\
Neutral-A & 25.9 & $-35.5$ \\
\hline
\end{tabular}
\caption{Wrong-anchor ablation on ARC. \atok{} is replaced by the correct option token (\emph{Correct-A}), a wrong option token (\emph{Wrong-A}), a random Gaussian vector (\emph{Random-A}), an all-zero vector (\emph{Zero-A}), or a neutral padding token (\emph{Neutral-A}). Non-semantic anchors cause large accuracy drops.}
\label{tab:wrong_a}
\end{table}

\begin{table}[t]
\centering
\resizebox{\columnwidth}{!}{
\begin{tabular}{l l c c c c}
\hline
Model & Dataset & SimLens & NoS & Drop & Layer \\
\hline
LLaMA-7B & ARC & 70.6 & 69.2 & $-1.4$ & 24 \\
LLaMA-7B & BoolQ & 84.9 & 81.9 & $-3.0$ & 16 \\
LLaMA-7B & HeadQA & 43.6 & 23.5 & $-20.1$ & 19 \\
Vicuna-7B & ARC & 75.4 & 75.0 & $-0.4$ & 26 \\
Vicuna-7B & BoolQ & 84.6 & 83.3 & $-1.3$ & 21 \\
Vicuna-7B & HeadQA & 51.2 & 13.2 & $-38.0$ & 20 \\
\hline
Mean & -- & 68.4 & 57.7 & $-10.7$ & 21 \\
\hline
\end{tabular}
}
\caption{SimLens-NoS ablation (removing \stok{}) at \SimLens{}'s best layer for each setting. SimLens / NoS columns are accuracies (\%), Drop is in pp, and Layer is the \SimLens{} best layer.}
\label{tab:simlens_nos}
\end{table}

Replacing Correct-A with Wrong-A causes a $13.2$\,pp drop, and the non-semantic Random-A/Zero-A/Neutral-A cause even larger drops ($35.5$--$36.5$\,pp), confirming that \atok{} is not a formatting placeholder but provides essential semantic anchoring for alignment with the candidate.

\subsection{Why the Start Token Matters: Global Condition}
\label{sec:ablation_s}
We further test \SimLens{}-NoS, which removes the start token from the reduced state. Removing the start token degrades all six settings with an average drop of $10.7$\,pp at \SimLens{}'s best layer; the effect is especially strong on HeadQA ($-20.1$\,pp on LLaMA-7B and $-38.0$\,pp on Vicuna-7B), where the answer space is more complex, and smallest on ARC ($-1.4$ and $-0.4$\,pp). The full per-setting results are reported in Table~\ref{tab:simlens_nos}. Together with the wrong-anchor results in Table~\ref{tab:wrong_a}, this indicates that robust \SimLens{} decoding depends on both semantic anchoring from the candidate answer token and global conditioning from the start token.

We read these ablations through two non-exclusive lenses: \emph{space alignment}---forwarding \stok{} and \atok{} through the upper layers may better align intermediate states with the output space---and \emph{extra non-linearity}---the extra forward step may recover transformations a linear map cannot express. We treat both as analysis rather than mechanistic proof; a supporting attention-allocation diagnostic is provided in Appendix~\ref{app:attention_diagnostic}.

\section{Conclusion}
We introduce \SimLens{}, a training-free readout that forwards a reduced two-token continuation through the model's upper layers for single-token decision tasks, and \LinearSimLens{}, a lightweight distilled monitor used by the hybrid early-exit framework \SimExit{}. Across different models, \SimLens{} improves fair-compute candidate readout over direct linear lenses, and \SimExit{} produces a modest but stable speed--accuracy frontier on the two models used end-to-end. Ablations indicate that the start token contributes global conditioning and the candidate answer token contributes semantic anchoring; we treat the underlying mechanism as analysis rather than a single hard claim.

\section{Limitations}
Our main experiments focus on single-token decision tasks in which candidate labels are supplied by the task format. While the same candidate-scoring formulation also underlies several production inference pipelines---such as RAG reranking, answer verification and routing, content moderation, and preference/reward scoring---our evaluation is limited to benchmark multiple-choice datasets, and we leave a study of \SimExit{} in these deployed settings to future work. We also do not position \SimLens{} as a general-purpose accelerator for unrestricted long-form generation. Within our current setting, extending \SimExit{} to multi-token autoregressive decoding requires skipped-layer KV-cache backfilling, which can erode or eliminate the ideal early-exit gain as generation length grows. A KV-consistent implementation can always fall back to full-depth decoding, so its normalized cost is bounded by the full-model cost rather than allowed to exceed it (Appendix~\ref{app:kv_backfill_proxy}). This accounting issue is isolated to the auxiliary multi-token limitation analysis and does not affect the main single-token readout, fair-compute, or threshold-sensitivity results. Finally, we compare against representative linear lenses but not every recent variant, and we leave a broader head-to-head against increasingly strong learned linear decoders to future work.

\section*{Acknowledgments} This work was supported by the Brain Science and Brain-like Intelligence Technology-National Science and Technology Major Project (grant no.~2021ZD0203701) to T.Y.

\bibliography{references}

\clearpage

\appendix

\section{Additional Motivation Visualizations}
\label{app:motivation}

\begin{figure}[t]
    \centering
    \includegraphics[width=0.95\linewidth]{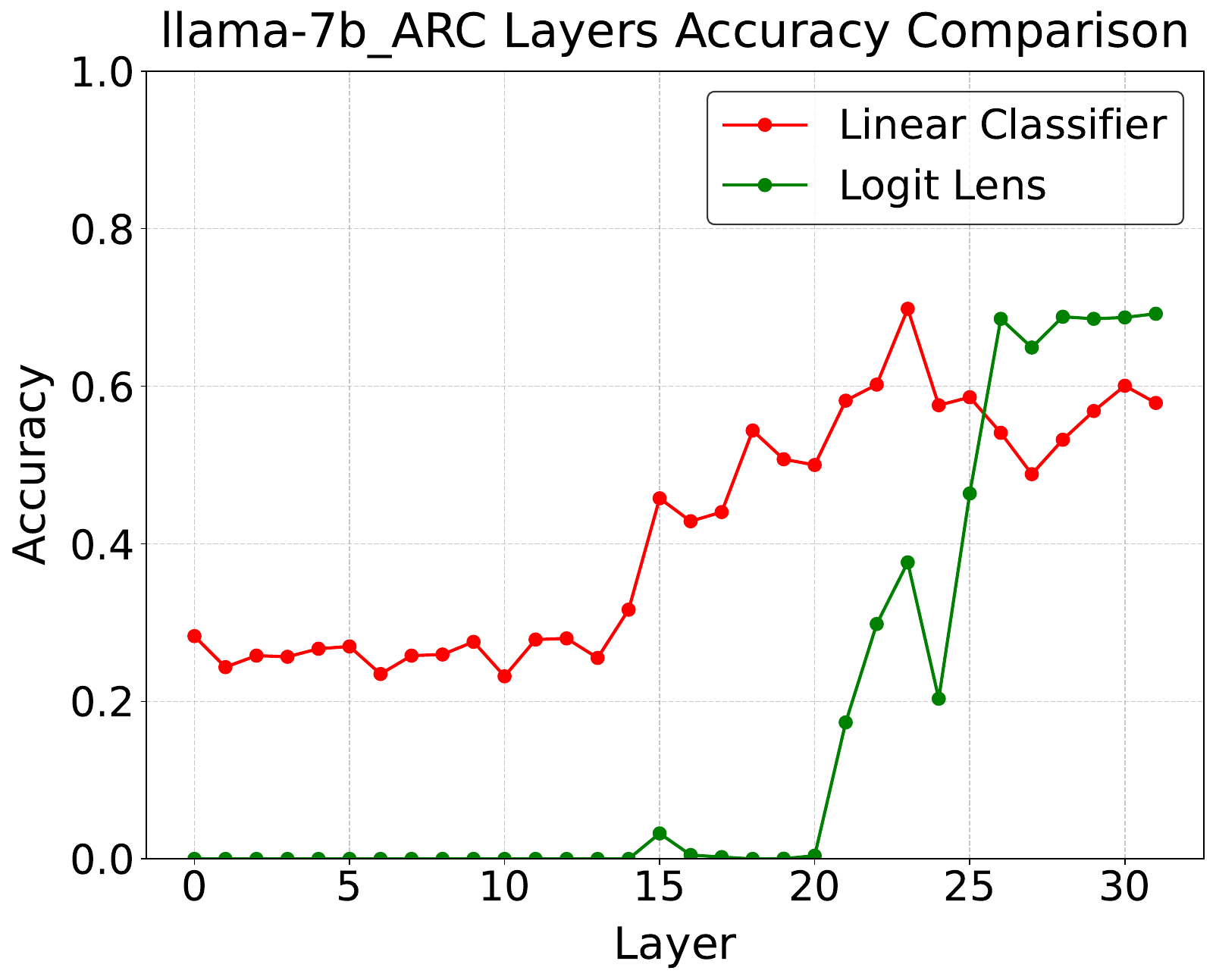}
    \caption{Motivation from representation-output mismatch on ARC. A linear probe recovers answer information much earlier than Logit Lens, indicating that useful task information emerges before final-layer alignment.}
    \label{fig:pca1}
\end{figure}

\begin{figure}[t]
    \centering
    \includegraphics[width=0.9\linewidth]{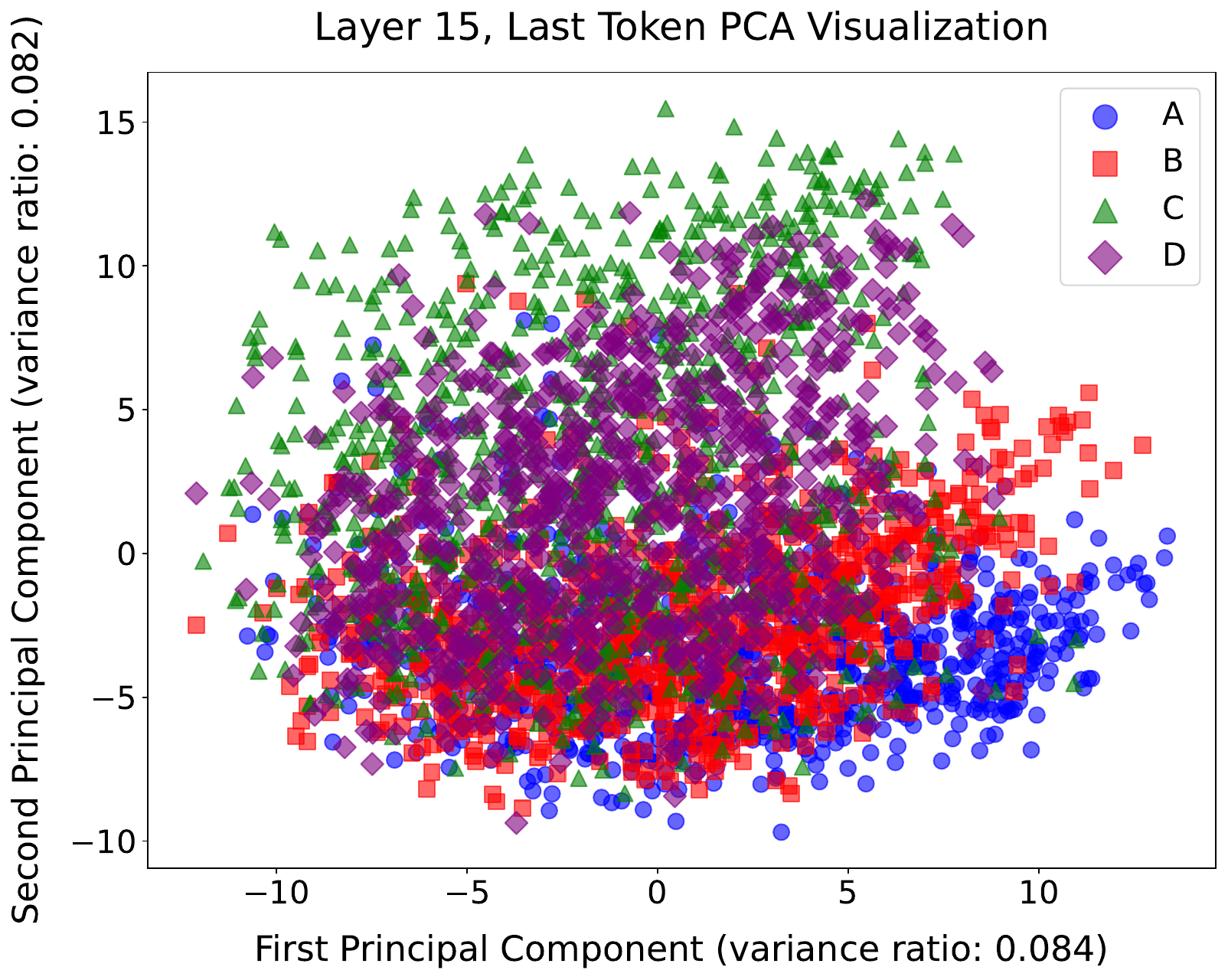}
    \caption{PCA snapshot of layer-15 last-token representations on ARC. Answer choices already form partially separable clusters, consistent with the early linear-probe advantage discussed in the main text.}
    \label{fig:pca_snapshot}
\end{figure}

Figure~\ref{fig:pca1} provides the main quantitative motivation: a linear probe recovers answer information much earlier than Logit Lens on ARC, indicating that useful task information is present before final-layer alignment. Figure~\ref{fig:pca_snapshot} adds a qualitative snapshot of this phenomenon at layer 15.

\section{Supplementary Experiments for SimLens}
\label{sec:other_simlens_performance}

\subsection{Cross-Dataset CE / Perplexity}
\label{app:ce_curves}

\begin{figure*}[t]
    \centering
    \includegraphics[width=\linewidth]{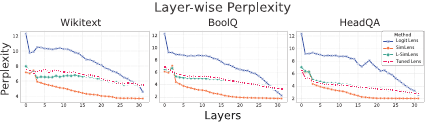}
    \caption{Layer-wise cross-entropy / perplexity on WikiText-103 (left), BoolQ (middle), and HeadQA (right). Linear SimLens and Tuned Lens mappings are trained once on ARC and directly evaluated on target datasets. Lower is better.}
    \label{fig:simlens_compare}
\end{figure*}

Figure~\ref{fig:simlens_compare} reports layer-wise cross-entropy / perplexity when Linear SimLens and Tuned Lens are trained once on ARC and directly evaluated on BoolQ, HeadQA, and WikiText-103 \citep{merity2016pointer}. SimLens (training-free) consistently attains the lowest CE across layers; Linear SimLens remains competitive at much lower monitoring overhead.

\subsection{Vicuna-7B Layer-wise Curves}

This appendix presents SimLens and Linear SimLens results on the Vicuna-7B model. All hyperparameters and evaluation protocols strictly follow those used for LLaMA-7B in the main text.

\begin{figure*}[htbp]
    \centering
    \includegraphics[width=\linewidth]{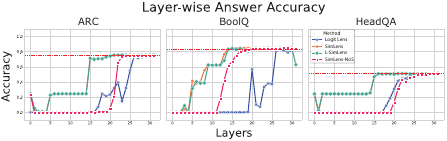}
    \caption{Vicuna-7B layer-wise decoding performance on ARC, BoolQ, and HeadQA for Logit Lens, SimLens, Linear SimLens, and SimLens-NoS.}
    \label{fig:vicuna_arc_simlens}
\end{figure*}

Figure~\ref{fig:vicuna_arc_simlens} shows SimLens-based decoding on Vicuna-7B for ARC, BoolQ, and HeadQA. Trends are consistent with LLaMA-7B: SimLens and Linear SimLens outperform Logit Lens, and removing the start token causes a clear performance drop. This supports SimLens' effectiveness on instruction-tuned models.

\begin{figure*}[htbp]
    \centering
    \includegraphics[width=\linewidth]{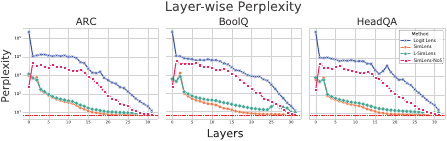}
    \caption{Layer-wise perplexity with vanilla LLaMA-7B across decoding methods. Lower perplexity indicates better readout quality.}
    \label{fig:simlens_ppl_compare}
\end{figure*}

All five backbones in the main layer-wise comparison use the same downstream data, code, and LoRA adaptation workflow \citep{hu2022lora}. For LLaMA-7B and Vicuna-7B, we additionally evaluate the untuned checkpoints. Although their overall perplexity is higher without fine-tuning, SimLens still outperforms the other readout baselines across most layers in Figure~\ref{fig:simlens_ppl_compare}.

\subsection{Linear SimLens Reproducibility}
\label{app:linear_simlens_repro}

We train one linear map per monitored layer with the underlying LLM---including its embedding and \texttt{lm\_head}---kept frozen, distilling from the full-vocab SimLens logits of Eq.~(\ref{eq:simlens}) on the ARC training split of each model/dataset setting. Optimization uses AdamW with a learning rate of $1\times10^{-3}$, batch size $4$, and distillation temperature $\tau=1.0$, run for $10$ epochs with checkpoint selection by validation loss; all runs use fp16 on a single A100. Exit thresholds are tuned on a held-out monitoring split.

\subsection{Position-id Implementation and Ablation}
\label{app:position_id}

For every SimLens experiment in the main text, the two-token state $[\stok{}, \atok{}]$ is forwarded with freshly indexed position ids $\tilde{\mathrm{pos}}=[0,1]$ rather than preserving \atok{}'s original prompt position. This follows the default behavior of our transformer wrapper when \texttt{inputs\_embeds} is passed without explicit \texttt{position\_ids}.

To check whether this choice is load-bearing, we ran a small ablation on 200 cached LLaMA-7B + ARC LoRA examples, comparing compact $[0, 1]$ against original $[0, L-1]$ positions for \atok{}. At the best SimLens layer (L21), the two schemes are identical at $79.5\%$ accuracy. On any layer where SimLens reaches above the random baseline (from L14 onwards), the two schemes agree within $1$\,pp; the only large differences appear at L0--L1, where SimLens is at chance under either scheme. We therefore retain compact positions throughout the main experiments. The per-layer numbers and the script that produces them are released with the paper.

\subsection{Layer-wise Readout Curves on Llama-3.1-8B and Qwen2.5-7B-Instruct}
\label{app:additional_backbones}

\begin{figure*}[htbp]
    \centering
    \includegraphics[width=0.48\linewidth]{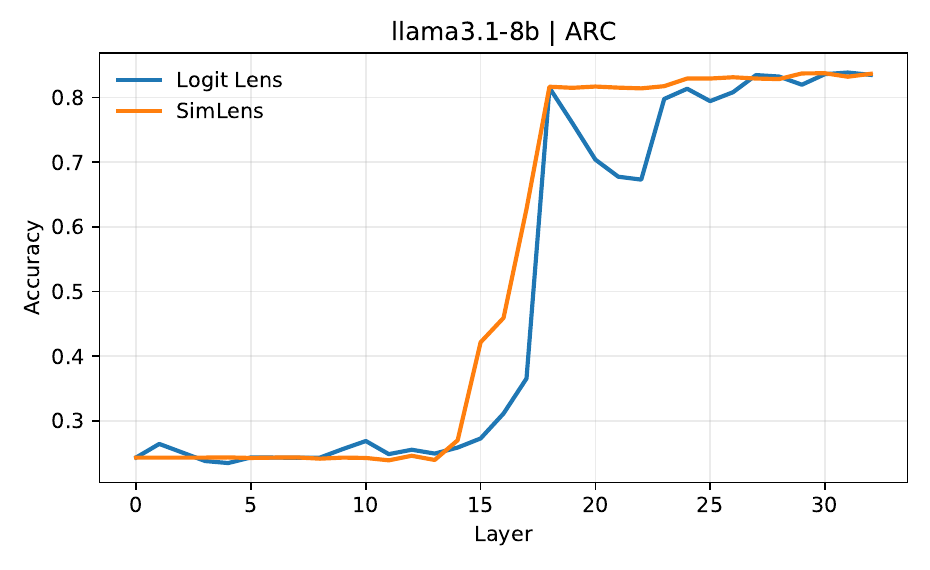}
    \hfill
    \includegraphics[width=0.48\linewidth]{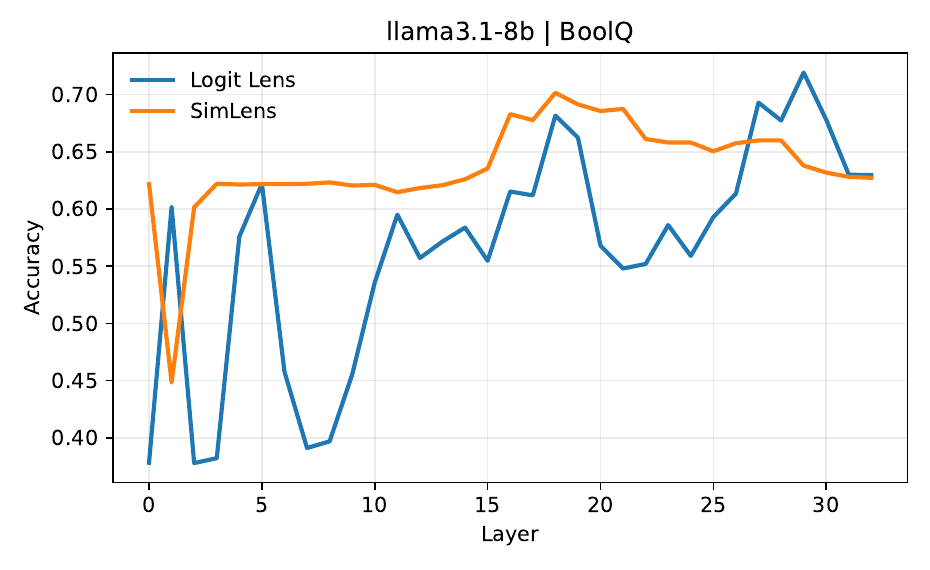}
    \\[0.5em]
    \includegraphics[width=0.48\linewidth]{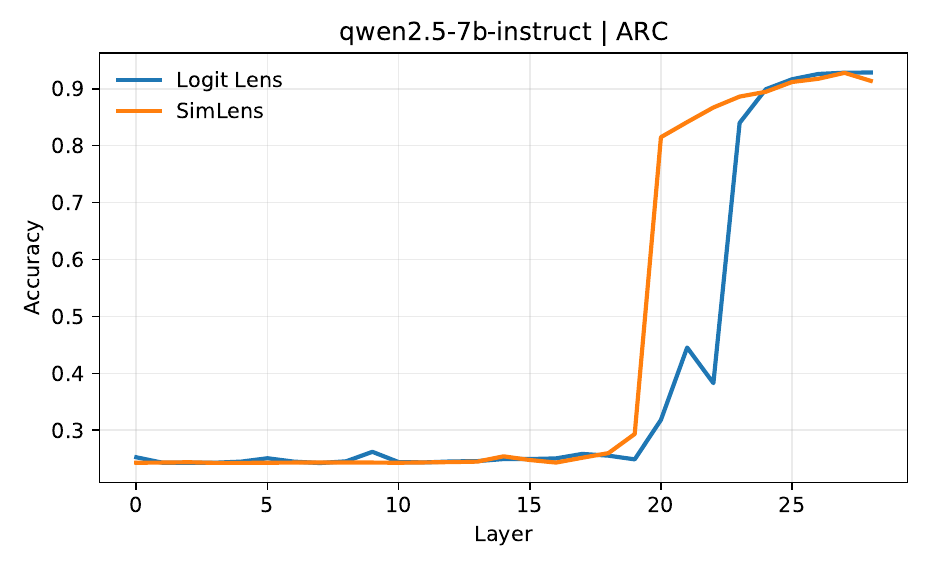}
    \hfill
    \includegraphics[width=0.48\linewidth]{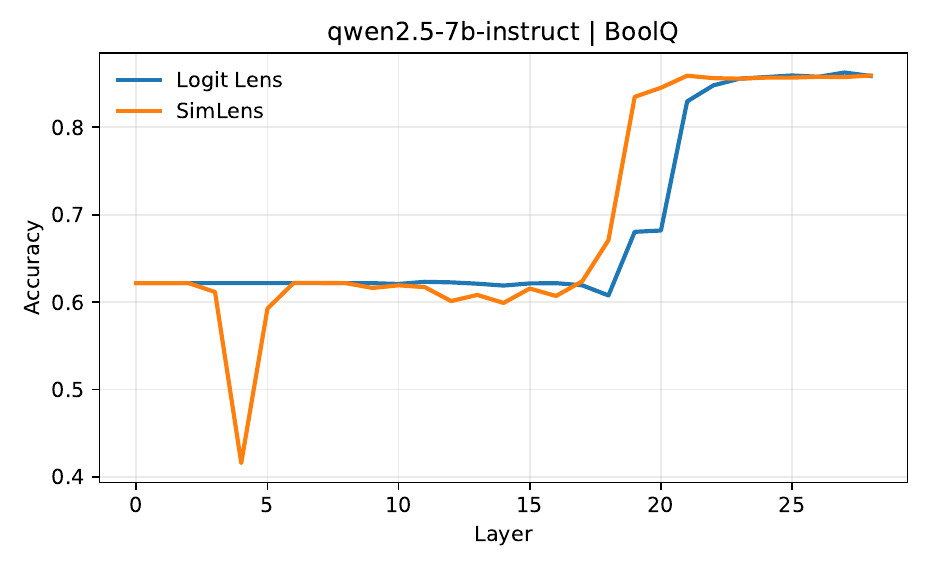}
    \caption{Layer-wise readout accuracy on Llama-3.1-8B and Qwen2.5-7B-Instruct under the candidate-scoring protocol. Top: Llama-3.1-8B on ARC (left) and BoolQ (right). Bottom: Qwen2.5-7B-Instruct on ARC (left) and BoolQ (right). The summary numbers underlying Table~\ref{tab:additional_backbones} are derived from these curves; the qualitative pattern (small best-layer gap against Logit Lens, larger advantage at earlier intermediate layers) is consistent across all four settings.}
    \label{fig:additional_backbones}
\end{figure*}

Figure~\ref{fig:additional_backbones} shows the full layer-wise Logit Lens vs.\ \SimLens{} curves on Llama-3.1-8B and Qwen2.5-7B-Instruct for ARC and BoolQ. These curves are the source of the summary numbers in Table~\ref{tab:additional_backbones}. The pattern of small best-layer gaps with a visible mid-depth advantage is consistent across all four settings.

\subsection{Attention Diagnostic}
\label{app:attention_diagnostic}

We report two complementary attention diagnostics. Both should be read as analysis evidence supporting the global-conditioning interpretation of \stok{}; neither is meant as an exit-confidence signal.

\paragraph{Attention from the readout position to prompt spans.}
On 500 LLaMA-7B + ARC examples we measure, at selected layers, the attention mass that the readout (final prompt) position allocates to four reconstructed input spans of the original full prompt: the start token, the question, the options, and the answer prefix. Selected layers are reported in Table~\ref{tab:attn_spans}. Attention to \stok{} grows quickly after the first few layers (e.g., $0.80$ at L4, $0.72$ at L28) while SimLens accuracy only rises sharply around L16--L24. The two are therefore not linearly tied.

\begin{table*}[t]
\centering
\begin{tabular}{r r r r r r}
\hline
Layer & \stok{} & Question & Options & Ans.\ prefix & SimLens acc (\%) \\
\hline
0 & 0.057 & 0.230 & 0.519 & 0.194 & 24.3 \\
4 & 0.803 & 0.010 & 0.049 & 0.138 & 24.3 \\
8 & 0.745 & 0.058 & 0.097 & 0.100 & 24.2 \\
12 & 0.509 & 0.125 & 0.237 & 0.129 & 24.3 \\
16 & 0.577 & 0.079 & 0.162 & 0.182 & 65.5 \\
20 & 0.440 & 0.064 & 0.203 & 0.293 & 70.3 \\
24 & 0.455 & 0.058 & 0.200 & 0.287 & 70.6 \\
28 & 0.719 & 0.010 & 0.049 & 0.223 & 70.4 \\
31 & 0.652 & 0.089 & 0.102 & 0.157 & 70.4 \\
\hline
\end{tabular}
\caption{Attention allocation from the readout position to reconstructed input spans of the original full ARC prompt, and matched-layer SimLens accuracy.}
\label{tab:attn_spans}
\end{table*}

\paragraph{Reduced-sequence \atok{}\,$\to$\,\stok{} attention.}
On Llama-3.1-8B ARC / BoolQ we additionally measure, inside the reduced two-token continuation, the layer-mean attention mass from \atok{} to \stok{}. Across $128$ ARC and $128$ BoolQ examples, this mass is $0.85$ on ARC and $0.85$ on BoolQ (means over layers, excluding L0). The correct-vs-incorrect gap is below $0.002$ at every layer, which is consistent with \stok{} acting as a stable global-conditioning anchor but rules out using this attention as an exit signal.

\section{Supplementary Experiments for SimExit}

\subsection{Detailed Threshold Tables}
\label{app:threshold_tables}

\begin{figure*}[t]
  \centering
  \includegraphics[width=\linewidth]{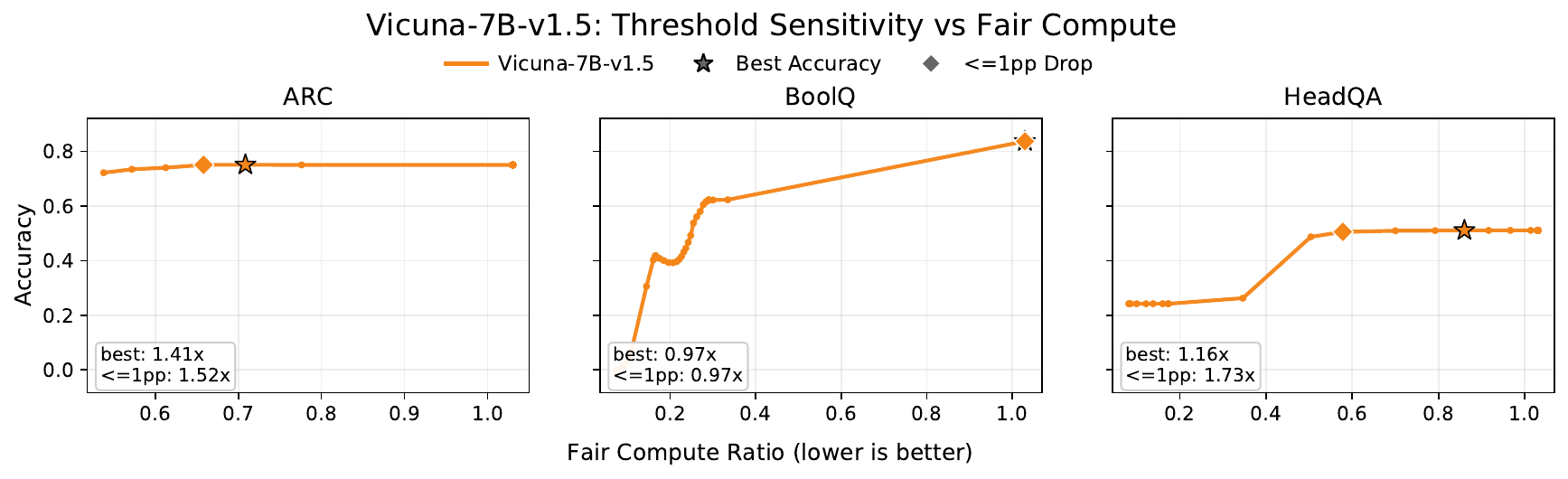}
  \caption{Accuracy vs. fair compute under threshold sweeping for Vicuna-7B. Fair compute includes the +2-token post-forward overhead after truncation. Star markers indicate best-accuracy operating points; diamond markers indicate operating points within 1pp accuracy drop.}
  \label{fig:threshold_fair_compute_vicuna}
\end{figure*}

Table~\ref{tab:threshold_sensitivity_extended} reports the exact operating points underlying the threshold-sensitivity summary in the main text. Figure~\ref{fig:threshold_fair_compute_vicuna} provides the corresponding fair-compute frontier for Vicuna-7B (the LLaMA-7B frontier is shown in the main text, Figure~\ref{fig:threshold_fair_compute_llama}).

\subsection{Multi-Token KV-Cache Backfill Proxy}
\label{app:kv_backfill_proxy}

\begin{figure}[t]
  \centering
  \includegraphics[width=\linewidth]{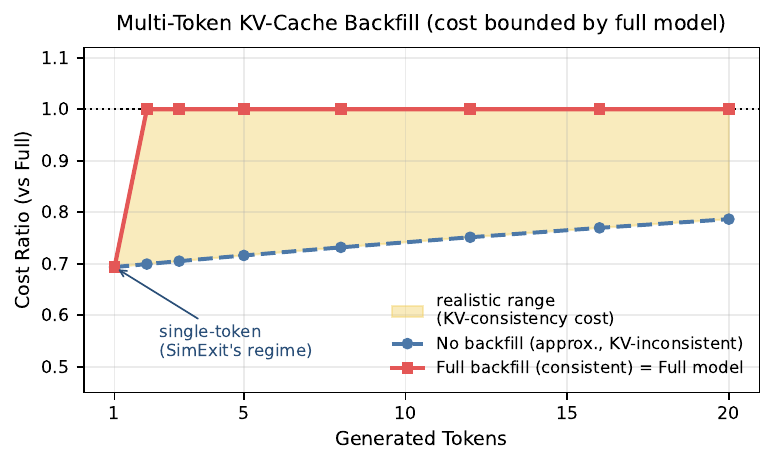}
  \caption{Multi-token limitation under KV-cache backfill assumptions. KV-consistent decoding is capped at the full-model cost, while the no-backfill curve is an approximate, KV-inconsistent lower-bound proxy. The shaded region marks the single-token setting evaluated by \SimExit{}.}
  \label{fig:limitation_backfill_scaling}
\end{figure}

For multi-token generation, an early-exit decoder must either backfill cache entries for skipped layers or revert to full-depth computation. The latter provides a strict fallback, so normalized cost is capped at $1.0$ (the full model) rather than allowed to exceed it. Backfilling can nevertheless consume the entire ideal saving as the output grows; our present experiments therefore do not establish acceleration for multi-token generation. Figure~\ref{fig:limitation_backfill_scaling} visualizes this corrected accounting: the no-backfill curve is only an approximate, KV-inconsistent lower-bound proxy, whereas KV-consistent decoding is bounded by falling back to the full model. This accounting issue is isolated to this auxiliary limitation analysis and does not affect the single-token layer-wise readout, fair-compute, or threshold-sensitivity results.

\subsection{Additional Performance on Other Datasets and Models}
\label{app:exit_perf}
This appendix reports SimExit results with LLaMA-7B and Vicuna-7B on ARC, BoolQ, and HeadQA using the same adaptive-threshold protocol as in the main text.

\begin{figure*}[htbp]
    \centering
    \includegraphics[width=\linewidth]{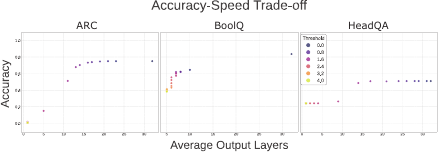}
    \caption{SimExit speed-accuracy curves on Vicuna-7B across ARC, BoolQ, and HeadQA under different entropy thresholds.}
    \label{fig:vicuna_arc_simexit}
\end{figure*}

Figure~\ref{fig:vicuna_arc_simexit} shows that SimExit consistently achieves competitive accuracy while enabling early termination across layers. These results confirm the general effectiveness of SimExit in a variety of QA tasks and models.

\subsection{Additional Generalization Performance on Other Datasets and Models}
This section expands on the generalization analysis, presenting full results where SimExit is applied to unseen datasets using Linear SimLens mappings trained on other datasets. Results are shown for both LLaMA-7B and Vicuna-7B across all dataset permutations (ARC, BoolQ, HeadQA).

\begin{figure*}[htbp]
    \centering
    \includegraphics[width=\linewidth]{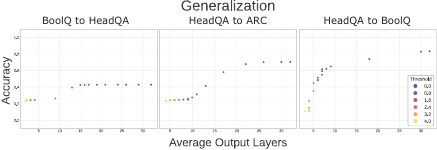}
    \caption{Cross-task generalization of SimExit with Linear SimLens on LLaMA-7B.}
    \label{fig:llama_arc_headqa_simexit_gene}
\end{figure*}

\begin{figure*}[htbp]
    \centering
    \includegraphics[width=\linewidth]{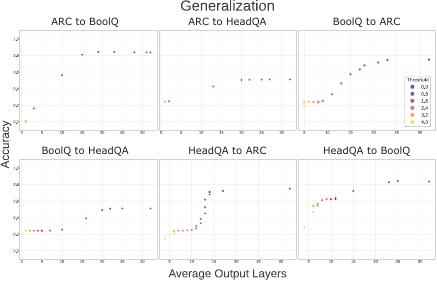}
    \caption{Cross-task generalization of SimExit with Linear SimLens on Vicuna-7B.}
    \label{fig:vicuna_arc_boolq_simexit_gene}
\end{figure*}

The performance curves in Figures~\ref{fig:llama_arc_headqa_simexit_gene} and \ref{fig:vicuna_arc_boolq_simexit_gene} demonstrate that SimExit generalizes well across different tasks.

\begin{table*}[t]
\centering
\resizebox{\textwidth}{!}{
\begin{tabular}{l l c c c c c c c c}
\hline
Model & Dataset & Best Th & Best Acc (\%) & Best Fair Cost & Best Speedup & Th @$\le$1pp Drop & Acc @$\le$1pp Drop (\%) & Fair Cost @$\le$1pp Drop & Speedup @$\le$1pp Drop \\
\hline
LLaMA-7B & ARC & 3.80 & 70.4 & 0.865 & 1.16$\times$ & 3.80 & 70.4 & 0.865 & 1.16$\times$ \\
LLaMA-7B & BoolQ & 7.40 & 83.2 & 0.951 & 1.05$\times$ & 7.20 & 83.1 & 0.744 & 1.34$\times$ \\
LLaMA-7B & HeadQA & 1.20 & 43.0 & 0.875 & 1.14$\times$ & 0.80 & 42.4 & 0.597 & 1.67$\times$ \\
Vicuna-7B & ARC & 0.80 & 75.1 & 0.709 & 1.41$\times$ & 0.60 & 75.0 & 0.658 & 1.52$\times$ \\
Vicuna-7B & BoolQ & 7.80 & 83.6 & 1.030 & 0.97$\times$ & 7.80 & 83.6 & 1.030 & 0.97$\times$ \\
Vicuna-7B & HeadQA & 2.80 & 51.1 & 0.860 & 1.16$\times$ & 2.20 & 50.5 & 0.578 & 1.73$\times$ \\
\hline
Mean & - & 3.97 & 67.7 & 0.882 & 1.149$\times$ & 3.73 & 67.5 & 0.745 & 1.399$\times$ \\
\hline
\end{tabular}
}
\caption{Threshold sensitivity of SimExit under fair-compute accounting, with explicit fair-cost fields at both operating points (best accuracy and within 1pp accuracy drop).}
\label{tab:threshold_sensitivity_extended}
\end{table*}

\end{document}